# D-ITAGS: A Dynamic Interleaved Approach to Resilient Task Allocation, Scheduling, and Motion Planning

Glen Neville, Sonia Chernova, Harish Ravichandar

*Abstract*—Complex, multi-task missions require the coordination of heterogeneous robots at multiple inter-connected levels, such as coalition formation, scheduling, and motion planning. This challenge is exacerbated by dynamic changes, such as sensor and actuator failures, communication loss, and unexpected delays. We introduce Dynamic Iterative Task Allocation Graph Search (D-ITAGS) to *simultaneously* address coalition formation, scheduling, and motion planning in *dynamic* settings involving heterogeneous teams. D-ITAGS achieves resilience via two key characteristics: i) interleaved execution, and ii) targeted repair. *Interleaved execution* enables an effective search for solutions at each layer while avoiding incompatibility with other layers. *Targeted repair* identifies and repairs parts of the existing solution impacted by a given disruption, while conserving the rest. In addition to algorithmic contributions, we derive accurate bounds on schedule suboptimality and provide insights into the inherent trade-off between time and resource optimality in these settings. Our experiments reveal that i) D-ITAGS is significantly faster than recomputation from scratch in dynamic settings, with little to no loss in solution quality, and ii) the theoretical bounds on optimality gap consistently hold in practice.

## I. INTRODUCTION

Heterogeneous multi-robot systems (MRS) bring together robots with complementary capabilities. They have been proved useful in domains as diverse as agriculture [1], assembly [2], and warehouse automation [3]. To achieve effective teaming in such complex domains, researchers have addressed challenging problems in coalition formation (*who*) [4], scheduling (*when*) [5], multi-robot motion planning (*how*) [3], and the combination of all these problems [6], [7].

Many algorithms developed for heterogeneous MRS coordination assume a *static* problem domain – one in which specifications and resources remain constant. However, real-world MRS do not enjoy the luxury of a predictable world, much less an unchanging one. Sensor and actuator failures, communication loss, and unexpected delays are all but a few examples. These events, even at the individual robot level, could cascade into catastrophic system-wide failures [8]. Note that robust task allocation methods [9], [10], while capable of effectively handling various forms of uncertainty, do not consider abrupt dynamic changes to the problem domain.

An obvious way to handle *dynamic* problems is to recompute the solution when unexpected events occur. However, as we demonstrate, this naïve approach is inefficient. Efficient approaches have been developed within the contexts of homogeneous robots [8], single-robot or decomposable tasks [11], and instantaneous task allocation [12]. However, we still lack approaches that can simultaneously handle task allocation, scheduling, and motion planning in dynamic settings involving multi-robot tasks and heterogeneous teams.

This work was supported by the Army Research Lab under Grants W911NF-17-2-0181 (DCIST CRA) and W911NF-20-2-0036

The authors are with the Institute for Robotics and Intelligent Machines, Georgia Institute of Technology, Atlanta, GA, USA {gneville, chernova, harish.ravichandar}@gatech.edu

This work introduces the *Dynamic Trait-Based Time Extended Task Allocation* problem. Our formulation can be seen as an instance of the well-known ST-MR-TA problem [13], with additional constraints to account for motion planning and dynamic changes. Specifically, we consider a variety of changes to the environment or the team (e.g., robot failures, task delays, etc.) that are unknown until after the fact.

To address the above, we develop *Dynamic Incremental Task Allocation Graph Search (D-ITAGS)*, an efficient algorithm to solve dynamic problems (see Fig. 1 for the architecture). D-ITAGS provides resilience against dynamic events due to two important characteristics: i) *interleaved executions* of individual modules, and ii) *targeted repair* of existing solutions.

First, leveraging our recent work [7], [14] which *interleaves* the execution of task allocation, scheduling, and motion planning, D-ITAGS effectively searches for solutions at each layer while ensuring compatibility with those at downstream layers. For instance, D-ITAGS will only consider allocations that do not violate scheduling constraints, and schedules with realizable motion plans. Indeed, we recently demonstrated that this interleaved approach is significantly more efficient than the often-used sequential approach [7], [14]. Compared to our prior methods, D-ITAGS includes a more efficient scheduler, and accounts for travel times to more tightly integrate scheduling and motion planning (see Sec. VII-A).

Second, we leverage the insight that many events do not render *all* computations performed for the existing solution invalid. For instance, when a robot is damaged, only allocations involving the damaged robot and the ones downstream are impacted. We develop a *targeted repair* mechanism that i) identifies and conserves parts of the solution that remain valid after an event, and ii) only recomputes parts that are now stale. Our approach can handle changes to i) robot capabilities, ii) task requirements, and iii) task duration, covering a wide spectrum of unexpected events. We demonstrate that targeted repair significantly decreases computation time compared to recomputing solutions from scratch, with little to no loss in solution quality (see Sec. VII-B).

In addition to the above computational and empirical benefits, we contribute theoretical insights into the operation of D-ITAGS. Specifically, we demonstrate that trait-based time-extended task allocation can be viewed as an inherent trade-off between *time optimality* (shortest makespan) and *resource optimality* (fewest allocations), and that one could traverse this trade-off spectrum by altering a single hyperparameter in our search heuristic. Leveraging this insight, we derive bounds on D-ITAGS' sub-optimality in terms of makespan under mild assumptions. These bounds provide guarantees on solution quality, and guide users in choosing hyperparameters. We also demonstrate that these bounds consistently hold in practice.

In summary, we contribute i) a formal definition of the dynamic trait-based time-extended task allocation problem, ii)

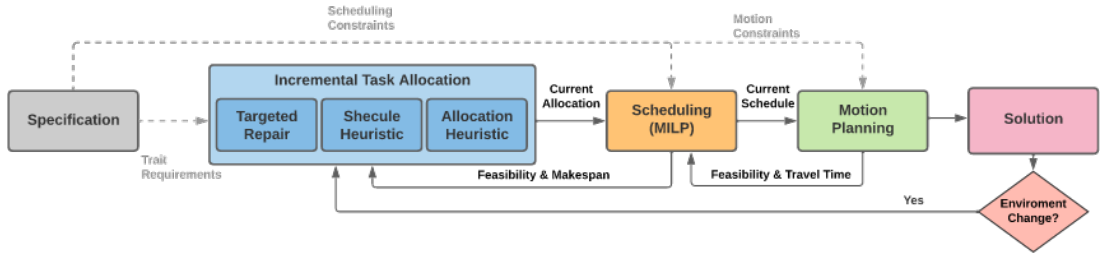

Figure 1: The proposed D-ITAGS algorithm leverages *targeted repair* and *interleaved* execution to simultaneously address coalition formation, scheduling, and motion planning in *dynamic* settings involving heterogeneous teams.

a resilient and efficient algorithm to handle dynamic changes, iii) theoretical insights into the inherent trade-off between time and resource optimality, and iv) performance guarantees.

## II. RELATED WORK

**ST-MR-TA methods:** While the multi-robot task allocation problem has many variants [13], [15], we limit our focus to single-task (ST) robots, multi-robot (MR) tasks, and time-extended (TA) allocation, as it closely relates to our work. ST-MR-TA problems require assigning coalitions of agents to tasks under temporal constraints. These constraints can take many forms, including precedence and ordering constraints, spatio-temporal constraints (e.g., travel time), and deadlines.

*Auction-based methods* to solve ST-MR-TA problems involve auctioning tasks to robots through a bidding process based on a utility function that combines the robot's (or the coalition's) ability to perform the task with any temporal constraints [16], [17]. Auctions have been shown to be highly effective, but typically either i) require multi-robot tasks to be decomposable into sub-tasks, each solvable by a single robot, or ii) assume that the ideal distribution of agents for each task is known (e.g., Task 1 requires one ground and one aerial robot). *Optimization-based methods* form another class of solutions that formulate the ST-MR-TA problem as a mixed-integer linear program (MILP) to optimize the overall makespan or a utility function [18]. However, these methods assume that some tasks can be left uncompleted [19] or require that all tasks be decomposable into single-agent tasks [20]. In contrast to auction-based and optimization-based approaches, our approach does not require task decompatibility and ensures the completion of all tasks.

Our approach to task allocation is most closely related to *trait-based methods* [4], [6], [14], [21], [22], which utilize a flexible modeling framework that encodes task requirements in terms of traits (e.g., Task 1 involves traveling at 10m/s while carrying a 50-lb payload). Each task is not limited to a specific set or number of agents. Instead, the focus is on finding a coalition of agents that *collectively* possess the required capabilities. However, most existing trait-based approaches are limited to ST-MR-IA problems that do not require scheduling [4], [14], [21], [22], with one notable exception [6]. Further, none of them can handle dynamic problems involving changes to the domain.

**Dynamic Task Allocation:** The approaches discussed above deal with *static* domains in which all aspects of the problem are known a priori and remain constant. Researchers have studied *dynamic* task allocation problems that consider unexpected events that occur during execution [23]. Similar to static problems, solutions to dynamic problems include game-theoretic methods [24], auction-based methods [11], and optimization methods [25]. However, these approaches do not solve the ST-MR-TA variant of task allocation. Further, existing approaches to dynamic problems inherit the limitations of their underlying methodology, and i) are limited to single-robot (SR) tasks [11], [23], ii) rely on decomposable tasks [20], iii) require specification of ideal agent distribution [23], or iv) entirely ignore scheduling and motion planning [12]. Complementary to approaches that explicitly consider dynamic events, *robust* task allocation methods [9], [10] attempt to find allocations that are robust to uncertainty and more likely to be valid even when information about the environment is uncertain. However, robust approaches require pre-specified models of uncertainty. In contrast, D-ITAGS leverages trait-based modeling and provides an efficient approach to handle unexpected changes in ST-MR-TA problems with heterogeneous robots.

## III. PROBLEM DESCRIPTION

We begin by formalizing the problem of *dynamic trait-based time-extended task allocation with spatial constraints*. We first present the static variant, which closely aligns with prior work [4], [6], [22], [26], and then introduce the dynamic variant.

Consider a team of $N$ heterogeneous robots, with the $i$th robot's capabilities described by a collection of traits $q^{(i)} = \left[ q_1^{(i)},\ q_2^{(i)},\ \cdots, q_U^{(i)} \right]$, where $q_u^{(i)} \in \mathbb{R}_{\geq 0}$ corresponds to the $u^{th}$ trait for the $i^{th}$ robot. We assign $q_u^{(i)} = 0$ when the $i^{th}$ robot does not possess the $u^{th}$ trait (e.g. firetrucks have a water capacity, but other robots may not). As such, the capabilities of the team can be defined by a **team trait matrix**:

$$\boldsymbol{Q} = \left[ q^{(1)\intercal},\ \cdots,\ q^{(N)\intercal} \right]^{\intercal} \in \mathbb{R}_+^{N \times U}$$

where $\boldsymbol{Q}_{iu}$ corresponds to the $i$th robot and $u$th trait.

We model the set of $M$ tasks that need to be completed as a **Task Network** $\mathcal{T}$: a directed graph $G = (\mathcal{E}, \mathcal{V})$, with vertices $\mathcal{V}$ representing a set of tasks $\{a_m\}_{m=1}^M$, and edges $\mathcal{E}$ represent relationships between any two tasks $a_i$ and $a_j$, such as a **precedence constraint** ($a_i \prec a_j$) requiring that Task $a_i$ be completed before the Task $a_j$ can begin (e.g., a fire must be put out before repairs can begin) and a **mutex constraint** ($a_i \neq a_j$) ensuring that $a_i$ and $a_j$ do not occur simultaneously (e.g. a robot cannot pick up two objects simultaneously). The team is required to complete all tasks, and robots can complete tasks individually or collaborate as a coalition, depending on the available resources.

Let the traits required to complete the $m$th task be denoted by $y^{(m)} = \left[y_1^{(m)}, y_2^{(m)}, \cdots, y_U^{(m)}\right]$, where $y_u^{(m)} \in \mathbb{R}_{\geq 0}$ is the amount of $u^{th}$ trait required for Task $m$. If the $u^{th}$ trait is not required by the $m^{th}$ task, we set $y_u^{(m)} = 0$. We can thus model the requirements of all tasks using the **desired trait matrix**:

$$\boldsymbol{Y}^* = \left[y^{(1)\intercal}, \cdots, y^{(N)\intercal}\right]^\intercal \in \mathbb{R}_+^{M \times U}$$

where $\boldsymbol{Y}^*_{mu}$ corresponds to the $m$th task and $u$th trait.

To execute Task $a_m$, a robot (or a coalition) requires a **collision-free paths** from its current configuration to the task's initial configuration $\mathcal{C}_I^m$, as well as from $\mathcal{C}_I^m$ to the task's final terminal configuration $\mathcal{C}_T^m$ (e.g., in a transport task, a robot needs collision-free paths to the package, as well as from the pickup to the dropoff). To compute such paths, a world model $W$ is provided which describes all of the static geometric information about the environment, including obstacles.

With the above definitions, we can define the **problem domain** using the tuple $\mathcal{D} = \langle \mathcal{T}, Q, \boldsymbol{Y}^*, I_c, L_T, W \rangle$, where $\mathcal{T}$ is the task network, $Q$ is the team trait matrix, $\boldsymbol{Y}^*$ is the desired trait matrix, $I_c$ and $L_T$ are respectively the sets of all initial and terminal configurations associated with tasks and $W$ is a description of the world state. Note that non-spatial tasks can be modeled by setting their initial and terminal configurations to be equal.

A solution to the problem specified by $\mathcal{D}$ consists of three components: *i)* an allocation of robots to tasks, *ii)* a task schedule, and *iii)* a set of associated motion plans.

We denote the allocation of robots to tasks using the **allocation matrix** $\mathbf{A} \in \mathcal{A}$ as follows

$$\mathbf{A} = \begin{bmatrix} o_{1,1} & \cdots & o_{1,n} \\ \vdots & \ddots & \vdots \\ o_{m,1} & \cdots & o_{m,n} \end{bmatrix}$$

where $o_{n,m} = 1$ if $n^{th}$ robot is assigned to the $m^{th}$ task. An allocation is considered valid when the *aggregated* traits of each coalition satisfy the trait requirements of the task to which it is assigned [4]. Formally, $\mathbf{A}$ is a **valid allocation** if and only if $\mathbf{A}Q$ is element-wise greater than or equal to $Y^*_\pi$. We denote the set of all valid by $\mathcal{N}_{sol}$.

Formally, we define the **solution** to the problem defined by $\mathcal{D}$ using the tuple $S = \langle \mathbf{A}, X, \sigma \rangle$ where $\mathbf{A}$ is a valid allocation, $X$ is a finite set of collision-free motion plans, and $\sigma$ is a schedule, represented by the set of start times for all tasks $\{s_i\}_{i=1}^M$, that respects all temporal constraints.

Note that the problem formulation from above assumes that all elements of the problem (e.g., tasks, requirements, robots, and world model) are *static*. As such, any valid solution $S$ to the problem defined in $\mathcal{D}$ would be rendered invalid when an unexpected change occurs (e.g., robot failures and introduction of new tasks). To capture *dynamic* environments with unforeseen changes, we define a new class of problems we call *Dynamic Trait-Based Time Extended Task Allocation*.

Formally, we model the dynamic problem domain as

$$\mathcal{D}_{\boldsymbol{k}} = \langle \mathcal{T}_k, Q_k, \boldsymbol{Y}^*_k, I_{ck}, L_{\mathcal{T}k}, W_k \rangle,$$

where the subscript $\boldsymbol{k}$ is used to denote the fact each domain definition is only valid for Iteration $k$ until an unexpected change redefines the problem domain to $\mathcal{D}_{k+1}$. Similarly, we denote a solution to the current problem definition using $S_{\boldsymbol{k}} = \langle \mathbf{A}_{\boldsymbol{k}}, X_{\boldsymbol{k}}, \boldsymbol{\sigma}_{\boldsymbol{k}} \rangle$ where $\mathbf{A}_k$ is a valid allocation, $X_k$ is a finite set of motion plans, and $\boldsymbol{\sigma}_k$ is a schedule for all tasks.

As we solve the dynamic trait-based time-extended task allocation problem, we are interested in two objectives: i) *time efficiency*: measured by the makespan $C(\sigma)$ of the Schedule $\sigma$, and ii) *resource sufficiency*: when aggregated traits $\mathbf{A}Q$ are element-wise greater than or equal to the requirements $Y^*$.

**Problem statement:** Given any new problem definition specified by $\mathcal{D}_{\boldsymbol{k}}$, compute the solution $S_{\boldsymbol{k}}$ by optimizing for time efficiency while ensuring resource sufficiency.

## IV. OVERVIEW OF APPROACH

To solve the dynamic trait-based time-extended task allocation problems with spatial constraints, as defined in Section III, we introduce our Dynamic Incremental Task Allocation Graph Search (D-ITAGS) algorithm. Though several approaches have been proposed for *instantaneous* trait-based task assignment [4], [21], [22], we are aware of only one approach – ITAGS [6] – that addresses the *time-extended* trait-based task allocation problem. Since D-ITAGS can be seen as a direct extension of ITAGS, we begin with a high-level summary of the similarities and differences, and provide details in Section V.

D-ITAGS shares two fundamental properties with ITAGS. First, D-ITAGS adopts a three-layer nested architecture from ITAGS, in which the processes of Task Allocation, Scheduling, and Motion Planning are interleaved. Second, D-ITAGS and ITAGS compute solutions incrementally by performing graph-based searches while leveraging heuristics.

D-ITAGS differs from and improves upon ITAGS in several ways. First, D-ITAGS constructs schedules by solving mixed-integer linear programs (MILPs) using GUROBI's branch-and-cut (B&C) solver, while ITAGS employs TABU search (TS). Note that the B&C solver we use can encode more temporal constraints and, unlike TS, provide bounds on suboptimality. While TS can produce schedules of comparable makespan, our experiments reveal that D-ITAGS is more efficient than ITAGS. Second, unlike ITAGS, D-ITAGS accounts for travel times between tasks while optimizing schedules by efficiently querying the motion planner for estimates. We demonstrate that these two differences significantly improve computational efficiency without sacrificing solution quality (see Section VII). Note that these improvements persist even when solving *static* time-extended trait-based task allocation.

The most significant difference is that D-ITAGS employs a *targeted repair* module when dynamic and unexpected changes render the existing solution invalid. In contrast, ITAGS was not designed to handle such unexpected events and thus would have to resort to recomputing a new solution from scratch.

## V. D-ITAGS

In this section, we discuss each module within D-ITAGS.

### A. Task Allocation

The task allocation layer of D-ITAGS uses a greedy best-first search through the task allocation space, $\mathcal{A}$. The task allocation space is modeled as a directed graph in which

Node $N$ denotes the allocation $\mathbf{A}_N$. The directed edge from a parent node $N_p$ to a child node $N$ represents the incremental allocation of a single robot to a particular task, with the root node containing no allocations. As such, all possible allocations exist within the full graph. Further, each node $N$ also contains a schedule $\sigma_N$ associated with its allocation $\mathbf{A}_N$. This graphical representation allows us to start from a node with no assignments and incrementally assign robots until we satisfy the trait requirements of all tasks.

Similar to ITAGS [6], we consider two heuristics to search the task allocation graph: *i) APR: Allocation Percentage Remaining* guides the search based on allocation quality, and *ii) NSQ: Normalized Schedule Quality* guides the search based on schedule quality. To balance the benefits of both heuristics, we use their convex combination, which we call *TETAQ: Time-Extended Task Allocation Quality*.

**APR** computes the percentage trait mismatch error as below

$$f_{apr}(\bar{N}) = \frac{||\max(E(\mathbf{A}_{\bar{N}}),\ 0)||_{1,1}}{||Y^*||_{1,1}} \quad (1)$$

where $\bar{N}$ is the node being evaluated and $\mathbf{A}_{\bar{N}}$ is its allocation, and $||\cdot||_{1,1}$ is the element-wise $l_1$ norm. The trait mismatch error $E(\mathbf{A}_{\bar{N}})$ is defined as $E(\mathbf{A}_{\bar{N}}) \triangleq Y^* - (\mathbf{A}_{\bar{N}} Q)$, where $\mathbf{A}_{\bar{N}} Q$ denotes the resources aggregated at each of the tasks given the allocation $\mathbf{A}_{\bar{N}}$. Note that any element of $\max(E(\mathbf{A}_{\bar{N}}),\ 0)$ will be zero if aggregated traits surpass the required traits of the corresponding task. As such, $f_{apr} \in [0, 1]$ quantifies the degree to which a given allocation meets the requirements.

**NSQ** measures the relative reduction in makespan as below

$$f_{nsq}(\bar{N}) = \frac{C(\sigma_{\bar{N}}) - C(\sigma_{LB})}{C(\sigma_{UB}) - C(\sigma_{LB})} \quad (2)$$

where $C(\cdot)$ returns the makespan of a given schedule, $\sigma_{\bar{N}}$ is the schedule associated with the node $\bar{N}$ being evaluated, $\sigma_{LB}$ is the estimated shortest schedule constructed by ignoring any constraints from allocation and motion planning, and $\sigma_{UB}$ is the longest schedule constructed by total ordering with the longest possible motion plans. While any upper bound on path length can be used, we use the sum of all edges in the task network $\mathcal{T}$ in our experiments. As such, $f_{nsq} \in [0, 1]$ quantifies the relative length of the schedule being evaluated.

**TETAQ** is a convex combination of APR and NSQ as below

$$f_{tetaq}(\bar{N}) = \alpha f_{apr}(\bar{N}) + (1-\alpha) f_{nsq}(\bar{N}) \quad (3)$$

where $\alpha \in [0, 1]$ is a user-specified parameter that controls the relative weighting of NSQ and APR heuristic.

### B. Scheduling and Motion Planning

D-ITAGS' scheduling layer checks the feasibility of scheduling a particular assignment and helps compute NSQ. We consider three different temporal constraints: precedence constraints, mutex constraints, and travel time constraints. Precedence constraints $\mathcal{P}$ ensure that one task happens before another (e.g., fire must be doused before repairs). Mutex constraints $\mathcal{M}$ ensure that two tasks do not happen simultaneously (e.g. a robot cannot pick up two objects simultaneously). Travel time constraints ensure that robots have sufficient time to travel between task sites (e.g., traveling to the location of fire before dousing). We formulate and solve the mixed-integer linear program as:

$$\begin{aligned}
\min_{\{s_i\}_{i=1}^M} \ & C \\
\text{s.t.} \ & C \geq s_i + d_i, \ \forall i = 1,..,M \\
& s_j \geq s_i + d_i + x_{ij}, \ \forall (i,j) \in \mathcal{P} \\
& s_i \geq x_i, \ \forall i = 1,..,M \\
& s_j \geq s_i + d_i + x_{ij} - M(1 - p_{ij}) \quad \forall (i,j) \in \mathcal{M}^R \\
& s_i \geq s_j + d_j + x_{ji} - M p_{ij} \quad \forall (i,j) \in \mathcal{M}^R
\end{aligned} \quad (4)$$

where $C$ is the makespan, $s_i$ and $d_i$ are the start time and duration of Task $a_i$, $x_{ij}$ is the time required to transition from $a_i$ to $a_j$, $p_{ij} = 1$ iff $a_i$ precedes $a_j$, $\beta \in \mathbb{R}_+$ is a large scalar, $\mathcal{P}$ and $\mathcal{M}$ are sets of integer pairs containing the lists of precedence and mutex constraints, with $\mathcal{M}^R = \mathcal{M} - \mathcal{P} \cap \mathcal{M}$ denoting mutex constraints with precedence constraints removed.

D-ITAGS constructs an initial schedule by estimating travel times based on the Euclidean distance between travel sites. As the search proceeds, the scheduling layer iteratively queries the motion planner to account for and accurately estimate travel times, until all motion plans required by the schedule are instantiated. Note that while our implementation uses the probabilistic roadmap planner [27], D-ITAGS is agnostic to the choice of motion planner.

D-ITAGS includes careful design choices that reduce the burden of motion planning placed on the scheduler. First, D-ITAGS memoizes all motion plans for future use. Second, it shares motion plans across robots with identical capabilities since their travel times between any two locations are likely similar. Further, our prior work has demonstrated that the interleaved approach adopted by D-ITAGS can handle significantly more tasks and robots than other existing approaches [6], [7].

In addition to constructing a valid schedule, the scheduling layer is also responsible for computing the bounds $\sigma_{UB}$ and $\sigma_{LB}$ used in the NSQ heuristic. For estimating the upper bound, we compute the worst-case makespan as follows

$$C(\sigma_{UB}) = \frac{2Mz}{w} + \sum_{m=1}^{M} d_m \quad (5)$$

where $z$ is the length of the longest possible path (e.g., the sum of all edges in the probabilistic roadmap) in $W$, and $w$ is the speed of the slowest robot. We set lower bound to be equal to the duration of the longest task: $C(\sigma_{LB}) = \max_m d_m$.

### C. Targeted Repair

When an unexpected event renders the current solution invalid, D-ITAGS efficiently recomputes a solution via targeted repair of the task allocation graph. D-ITAGS can specifically handle three categories of unexpected changes: i) changes in robots' capabilities, ii) changes in task requirements, iii) changes in task duration. Note that these three classes of changes encompass a wide variety of events, such as loss or reinforcement of robots, unexpected additional tasks, partial loss of robot capabilities, and unforeseen delays due to environmental conditions. Our only assumption is that any of the above specified events can be detected and identified using other techniques (e.g., [12]). In Section VII, we demonstrate D-ITAGS' ability to handle eight distinct event types.

To enable efficient repair, D-ITAGS leverages the inherent structure of the task allocation graph to identify and repair only a subset of the nodes. For example, when a robot's sensor is damaged or lost, D-ITAGS will identify nodes in the graph that allocated the damaged robot and recompute their APR, ignoring other nodes that will not contribute to a new solution.

D-ITAGS makes changes to three types of nodes: i) *open set*: collection of unexpanded nodes, ii) the *closed set*: collection of expanded nodes, and the iii) *pruned set*: collection of nodes deemed infeasible. We provide a detailed flowchart describing D-ITAGS' targeted repair mechanism in Fig. 2. D-ITAGS begins by adding the (now invalid) existing solution to the open set. Then, it incrementally checks for five specific changes and makes modifications as described below:

- *Agent or Task Loss*: When an agent or a task is lost, D-ITAGS removes nodes that utilize the lost agent or those that involve the lost task. When a task is lost in particular, D-ITAGS checks to find nodes in the open and closed set that might satisfy the requirements ($APR = 0$) after the task loss, and then continues the search with only the newly-identified solutions.
- *Reduced Traits or Increased Requirements*: When agent capabilities decrease or task requirements increase, the nodes in the closed and pruned sets remain infeasible ($APR < 0$). As such, D-ITAGS ignores them and only updates the APR of the nodes in the open set.
- *Increased Traits or Reduced Requirements*: When agent capabilities increase or task requirements decrease, D-ITAGS updates the APR of the nodes in the open set and checks if any of nodes in the closed and pruned sets have now become viable ($APR = 0$) after the increase (decrease) in capabilities (requirements).
- *Changed Duration*: When task duration or travel times change, APR remains unchanged. As such, D-ITAGS updates the NSQ of the nodes in the open set and ignores all other nodes as they remain infeasible ($APR < 0$).
- *New Agent*: When a new agent becomes available, D-ITAGS adds a new node as a child of the root node, and in turn appends the open set.

Leveraging the above insights, D-ITAGS avoids unnecessary recomputations when responding to dynamic events.

## VI. THEORETICAL ANALYSES

To better understand D-ITAGS' performance, we analyze the effect of $\alpha$ – the user-specified parameter that determines the relative importance of our two heuristics – on the optimality of the obtained solution. We consider two notions of optimality: i) time optimality (shortest makespan), and resource optimality (fewest assignments). We demonstrate that the choice of $\alpha$ determines the trade-off between the two notions of optimality, with each extreme value, $\alpha = 0$ or $\alpha = 1$, respectively guaranteeing time or resource optimality.

### A. Analysis of Time Optimality

We derive strict bounds on the time optimality gap of solutions generated by D-ITAGS as measured by makespan.

**Theorem 1.** *For a given trait-based time-extended task assignment problem, let $C(\sigma^*)$ be the optimal makespan and $C(\hat{\sigma})$ be the makespan of the solution generated by D-ITAGS. If $\alpha < 0.5$ in Eq. (3), then*

$$C(\hat{\sigma}) - C(\sigma^*) \leq \frac{\alpha}{1-\alpha}(C(\sigma_{UB}) - C(\sigma_{LB})) \quad (6)$$

*where $C(\sigma_{LB})$ and $C(\sigma_{UB})$ are estimated lower and upper bounds, respectively, on the makespan of any valid solution.*

*Proof.* Since any expansion of a parent node represents the addition of an assignment, any given node $N$ is guaranteed to have more agents assigned than its parent $N_p$. This observation, when combined with the fact that adding assignments can never reduce the makespan (as adding assignments can only introduce new constraints to the MILP), yields

$$f_{nsq}(N) \geq f_{nsq}(N_p) \quad (7)$$

Consequently, we can infer that the NSQ value of all nodes in the unopened set $\mathcal{U} \subseteq \mathcal{N}$ of a D-ITAGS graph is lower bounded by that of their respective predecessors in the opened set $\mathcal{O} \subseteq \mathcal{N}$. As such, the smallest NSQ value in the unopened set must be greater than that in the opened set:

$$\min_{N \in \mathcal{U}} f_{nsq}(N) \geq \min_{N \in \mathcal{O}} f_{nsq}(N) \quad (8)$$

Irrespective of the location of the node $N^*$ with optimal makespan $C(\sigma^*)$, the inequality in (8) implies that

$$C(\sigma^*) \geq \min_{N \in \mathcal{O}} C(\sigma_N) \quad (9)$$

As we require any valid solution to satisfy all trait requirements, the solution node ($\hat{N}$) will have an APR value of zero. Thus, the TETAQ heuristic (defined in (3)) of $\hat{N}$ is given by

$$f_{tetaq}(\hat{N}) = (1 - \alpha)f_{nsq}(\hat{N}) \quad (10)$$

Given the relationship in (8) and the fact that D-ITAGS selects a solution from the open set based on a best-first search, the TETAQ value of the solution node can be bounded as follows

$$f_{tetaq}(\hat{N}) \leq f_{tetaq}(N), \forall N \in \mathcal{O} \quad (11)$$

Expanding the definition of TETAQ and using (10) leads to

$$(1-\alpha)\frac{C(\hat{\sigma}) - C(\sigma_{LB})}{C(\sigma_{UB}) - C(\sigma_{LB})} \leq \\ \alpha f_{apr}(N) + (1-\alpha)\frac{C(\sigma_N) - C(\sigma_{LB})}{C(\sigma_{UB}) - C(\sigma_{LB})}, \forall N \in \mathcal{O} \quad (12)$$

Using the inequality in (8), the bound in (9), and the fact that $f_{apr}(\cdot) \leq 1$, we rewrite the above equation as

$$(1-\alpha)\frac{C(\hat{\sigma}) - C(\sigma_{LB})}{C(\sigma_{UB}) - C(\sigma_{LB})} \leq \alpha + (1-\alpha)\frac{C(\sigma^*) - C(\sigma_{LB})}{C(\sigma_{UB}) - C(\sigma_{LB})}$$

By simplifying and cancelling equivalent terms, we get

$$(1-\alpha)(C(\hat{\sigma}) - C(\sigma_{LB})) \leq \\ \alpha(C(\sigma_{UB}) - C(\sigma_{LB})) + (1-\alpha)(C(\sigma^*) - C(\sigma_{LB})) \quad (13)$$

On rearranging the terms, we arrive at the bound in (6). □

Note that the above result on time optimality gap is sensible only when $\alpha < 0.5$. When $\alpha \geq 0.5$, the bound in (6)

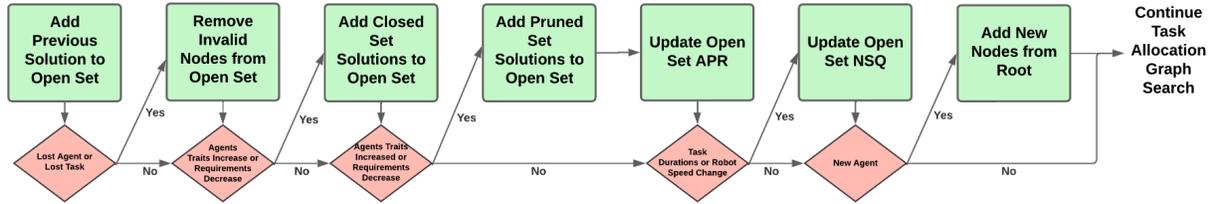

Figure 2: Flow chart illustrating D-ITAGS' targeted repair mechanism when responding to different unexpected events.

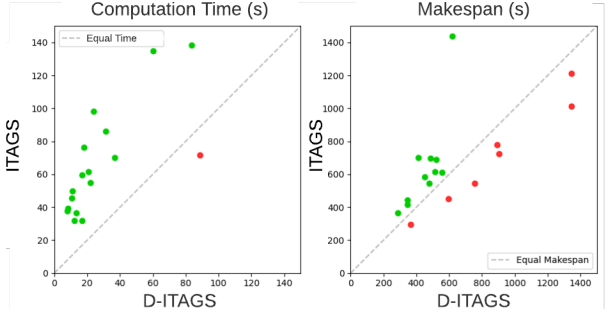

Figure 3: D-ITAGS is considerably faster than ITAGS (left), without sacrificing solution quality (right). Runs in which D-ITAGS is better (worse) than ITAGS are shown in green (red).

loses significance as it grows beyond the maximum difference possible difference in makespan ($C(\sigma_{UB})$ - $C(\sigma_{LB})$).

The bound presented in (6) can be tightened after the execution of the D-ITAGS, facilitating post-hoc performance analyses. Specifically, instead of bounding $f_{apr}(N), \forall N \in \mathcal{O}$ by 1, we can compute the exact minimum ($\min_{N \in \mathcal{O}} f_{apr}(N)$). Following similar algebraic manipulations as in the proof above, a tighter bound can then be derived as

$$C(\hat{\sigma}) - C(\sigma^*) \leq \frac{\alpha}{1-\alpha}(C(\sigma_{UB}) - C(\sigma_{LB})) \min_{N \in \mathcal{O}} f_{apr}(N) \quad (14)$$

### B. Analysis of Resource Optimality

Below, we show that we achieve resource optimality (i.e., the fewest number of assignments) when $\alpha = 1$.

**Theorem 2.** *Let $R(A) = \|A\|_1$ denote the total number of assignments in $A$, and $A_{N_{A^*}}$ be the allocation with the fewest assignments. Then, when $\alpha = 1$,*

$$R(A_{\hat{N}}) - R(A_{N_{A^*}}) = 0 \quad (15)$$

*Proof.* Since any expansion of a parent node represents the addition of an assignment, any given node $N$ is guaranteed to have more agents assigned than its parent $N_p$. As additional agents can never increase APR, we have

$$f_{apr}(N) < f_{apr}(N_p), \quad \forall N \in \mathcal{N} \quad (16)$$

The above inequality and the fact that D-ITAGS will only consider APR (and ignore NSQ) when $\alpha = 1$ suggest that D-ITAGS will begin at the root node and continue along the same branch until finding a solution, expanding only the nodes that decrease APR by the largest amount. As such, D-ITAGS will take the shortest route to the solution when $\alpha = 1$, leading to the fewest assignments (i.e., resource optimality). □

We do not include the bound on resource suboptimality in the interest of space and due to the fact that true resource optimality gap can be efficiently computed post-hoc since D-ITAGS takes the least amount of time to compute the resource optimal solution ($\alpha = 1$), in stark contrast to the time-optimal solution ($\alpha = 0$) required to compute the time optimality gap.

## VII. EVALUATION

We evaluated D-ITAGS using three sets of experiments in a simulated emergency response domain [6], [26], [28]–[30] in which a team of robots must rescue survivors, deliver medicines, douse fires, and rebuild damaged buildings. We generated problems from this domain by varying the number of robots between 8 and 16, the number of tasks between 20 and 40, and the location of tasks. In all experiments, we used maps from the Robocup Rescue Competition [30].

### A. Comparison to Existing Task Allocation Algorithms

In the first set of experiments, we analyzed how D-ITAGS performed on 20 *static* problems in our survivor domain relative to the ITAGS [6]. We chose to compare our approach with ITAGS as i) it has been shown to perform better than other state-of-the-art time-extended task allocation algorithms, and ii) ITAGS' trait-based task allocation inspired our approach.

We compared the performance of D-ITAGS and ITAGS in terms of computation time and solution makespan (see Fig. 3). As can be seen, D-ITAGS is capable of producing high-quality solutions on par with ITAGS while requiring far less computation time. The superior computational efficiency of D-ITAGS demonstrates that the D-ITAGS' branch-and-cut method is considerably more efficient than ITAGS' Tabu search. Given these observations and the fact that ITAGS has been shown to outperform state-of-the-art algorithms for ST-MR-TA [6], [7], we can conclude that D-ITAGS offers state-of-the-art computational efficiency without sacrificing quality.

### B. Performance on Dynamic Reallocation

We evaluated D-ITAGS on *dynamic* task allocation problems in which unexpected events can alter the problem domain at any point of execution, and compared it against the default strategy of existing approaches: running the task allocation algorithm from scratch given the updated problem domain. Note that the baseline used in these experiments is identical to D-ITAGS, except for missing the crucial targeted repair module. As such, any observed improvements can be attributed to the repair module and not to other improvements introduced by D-ITAGS (e.g., the MILP-based scheduler). We created a set of 500 dynamic repair problems in our survivor domain, and separated them into ten groups of 50 problems each. We measured the performance of both D-ITAGS and simple reallocation in terms of computation time and solution makespan.

Across a wide variety of dynamic conditions, we found that D-ITAGS produced high-quality solutions on par with reallocation from scratch (see Fig. 5), but required significantly less computation time (see Fig. 4). These improvements in computation efficiency are likely due to D-ITAGS' ability to identify and repair only the impacted nodes, while reusing the other nodes. We also found that D-ITAGS is particularly faster on problems that affect the allocations of nodes (e.g., changes to agents traits or the loss/addition of agents) as such changes do not require D-ITAGS to recompute the schedules of the affected nodes, saving expensive optimizations. Even for problems that require D-ITAGS to recompute schedules (e.g., task duration changes and lost/gained tasks), D-ITAGS performs significantly better than naive reallocation as it reuses cached motion plans and allocations from the existing solution that remain valid. It is important to note that D-ITAGS' solution quality could be worse than simple reallocation when a new agent is gained unexpectedly. This is caused by the fact that D-ITAGS' targeted repair favors efficiency by reusing valid existing nodes, even if they do not utilize the newly available agent. While the baseline benefits from the new agent as it reallocates, it takes longer to compute a solution.

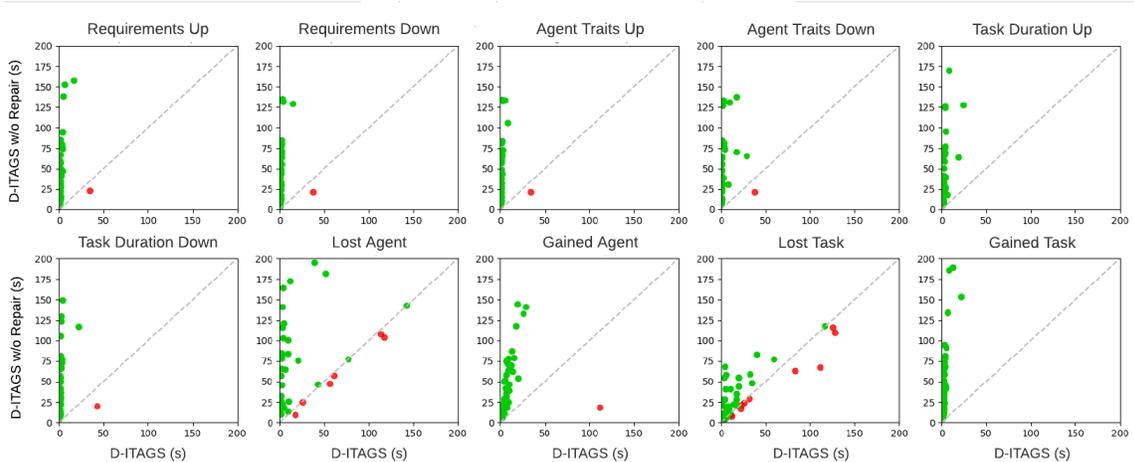

Figure 4: After unexpected changes, D-ITAGS (targeted repair) is significantly faster than recomputing solutions from scratch. Green (Red) dots indicate instances in which targeted repair performs better (worse) than complete reallocation.

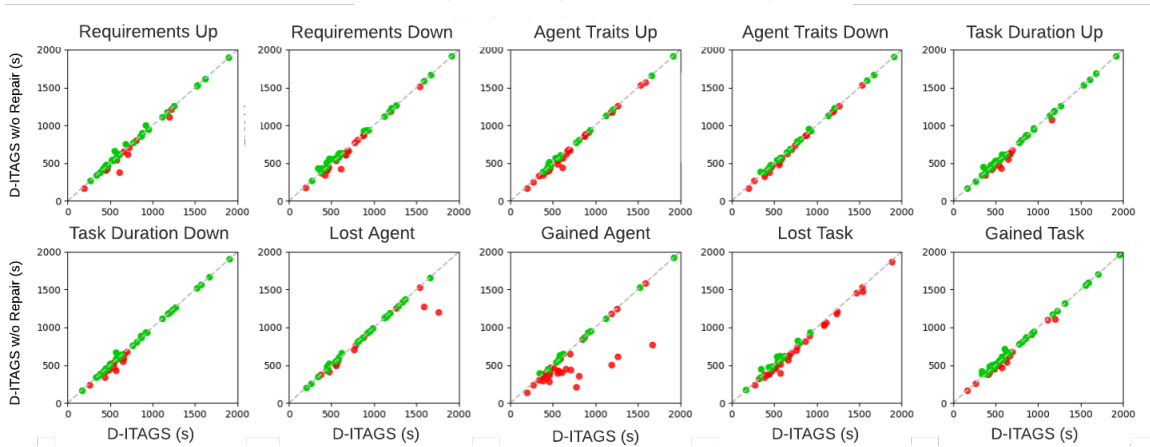

Figure 5: D-ITAGS (targeted repair) generates solutions of quality (makespan) similar to recomputing solutions from scratch. Green (Red) dots indicate instances in which targeted repair performs better (worse) than complete reallocation.

### C. Validation of Makespan Guarantees

In our final experiment, we empirically examined the validity of our theoretical guarantees on makespan from Sec. VI-A.

We created a set of 35 problems in our survivor domain, each of which was solved multiple times while varying $\alpha$ between $[0, 0.5]$. For every combination of problem and $\alpha$ value, we computed the actual normalized time optimality gap and the corresponding normalized theoretical bound. As can be seen from Fig. 6, the time optimality gaps consistently respect the theoretical bound across all values of $\alpha$. As expected, the values $\alpha = 0$ (ignore APR) and $\alpha = 1$ (ignore NSQ) respectively result in the shortest and longest schedules.

### VIII. CONCLUSIONS

We introduced D-ITAGS, an algorithm for task allocation in dynamic environments involving heterogeneous robots. D-ITAGS achieves efficient resilience by estimating and conserving portions of the existing solution that remain unaffected by the change. We showed that D-ITAGS trades-off between time and resource optimality and comes with theoretical guarantees on performance. Our detailed experiments conclusively demonstrate the effectiveness of D-ITAGS and its relative computational benefits over existing state-of-the-art task algorithms that resort to complete reallocation. A notable limitation of our work is that D-ITAGS assumes that unexpected events can be detected and identified. While such approaches are currently being developed (e.g., [12]), we still require tools that can detect events and solve the credit assignment problem.

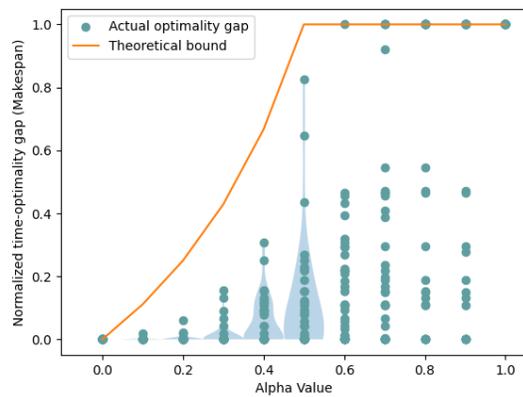

Figure 6: The theoretical bound consistently holds for varying values of $\alpha$. A value of zero for a normalized time-optimality gap represents an optimal schedule, and a value of one represents the longest schedule seen during allocation.

While D-ITAGS provides an efficient mechanism to solve dynamic ST-MR-TA problems, there are opportunities for improvement. First, the relative benefits of repair over complete reallocation are yet to be fully characterized. While D-ITAGS always chooses to repair, it might sometimes be beneficial to reallocate from scratch to better leverage positive changes despite the additional computation cost (e.g., our results show that reallocation might produce better-quality solutions when new agents become available, albeit at a significantly higher computational burden). Second, not all events *require* repair. For instance, when requirements reduce, we could continue using the current solution as it would remain valid despite becoming inefficient. In such circumstances, the benefits of repair must be weighed against its computational cost.